\documentclass[12pt]{article}

\usepackage{amsmath,amssymb,amsfonts,mathtools,amsthm}
\usepackage{color}

\newtheorem{theorem}{Theorem}[section]

\numberwithin{equation}{section}

\newcommand{\R}{\mathcal{R}}
\newcommand{\F}{\mathcal{F}}

\newcommand{\SA}{\mathcal{S}}
\newcommand{\PA}{\mathcal{P}}

\newcommand{\M}{\mathcal{M}}
\newcommand{\N}{\mathcal{N}}
\newcommand{\LA}{\mathcal{L}}

\newcommand{\T}{\mathcal{T}}
\newcommand{\0}{\textbf{0}}

\begin{document}

\begin{center}
\textbf{STOCHASTIC GRADIENT DESCENT WITH\\ \ \\  DISCONTINUITY ACROSS A MANIFOLD}
\end{center}

\bigskip

\begin{center}
Vivek S.\ Borkar\footnote{ Work supported by Google Research Asia and a National Science Chair from the Government of India} \\
Department of Electrical Engineering, \\
Indian Institute of Technology Bombay, \\
Powai, Mumbai 400076, INDIA.\\
(E-mail: borkar.vs@gmail.com)
\end{center}

\noindent \textbf{Abstract:} Stochastic gradient descent for a loss function discontinuous across lower dimensional manifolds is analyzed by studying its differential equation limit.  \\


\section{Introduction}

Stochastic Gradient Descent (SGD) has been the main workhorse for empirical risk minimization in machine learning. Both the basic SGD and its many variants have been extensively studiesd in the literature, both theoretically and empirically. Here we consider an important aspect of SGD which has not attracted adequate attention despite its important role in applications. This is the problem of discontinuities. We focus on a particular aspect here, viz, when the gradient of the function being minimized is discontinuous across a lower dimensional embedded manifold. We are interested in the case where the gradient vector field is transversal to and pointing into the manifold. This is a special case of stochastic approximation with discontinuous dynamics analyzed in literature \cite{Buck}, \cite{Faure1}, \cite{Faure2}, \cite{Yin}, \cite{Roth} and we do exploit the additional structure. Such dynamics can arise, e.g., when the loss function is not differentiable everywhere (e.g., when it is the point-wise maximum of finitely many smooth functions) or because the neural network architecture being designed has components whose input-output maps are not differentiable across a well structured boundary.\\

The next section formulates the problem in the above context. Section 3 then presents the basic results about ODE (for `\textit{Ordinary Differential Equations}') with vector field discontinuous across a lower dimensional manifold. Section 4 then states and proves our main results. Section 5 analyzes the time evolution.\\

We briefly recall here some related literature. SGD is a special case of a class of algorithms known as stochastic approximation and the approach we follow is the so called ODE approach which analyzes the algorithm using a limiting ODE (for `\textit{Ordinary Differential Equations}') associated with it. This approach was developed in the seventies through the works of Derevitskii and Fradkov \cite{Der}, Ljung \cite{Ljung}, Meerkov \cite{Meer}. Improved modern treatments with subsequent developments appear \cite{Benaim0}, \cite{Benv}, \cite{BorkarBook}, \cite{Duflo}. \\

We conclude this section with a summary of key notation. We denote by $\R^d$ the $d$-dimensional euclidean space and by $C([0,T]; \R^d)$ for $0 < T < \infty$ the Banach space of continuous maps $[0, T] \mapsto \R^d$ with the supremum norm $\|x(\cdot)\| := \max_{t\in [0,T]}\|x(t)\|$. $C([0,\infty);\R^d)$ will denote the Frechet space of continuous maps $[0, \infty) \mapsto \R^d$ with the corresponding inductive topology, also called the topology of uniform convergence on compacts. Thus $x_n(\cdot) \to x_\infty(\cdot)$ in $C([0,\infty);\R^d)$ if and only if 
$$x_n(\cdot)\Big|_{[0,T]} \to x_\infty(\cdot)\Big|_{[0,T]}$$ 
in $C([0,T];\R^d)$ 
for all $0 < T < \infty$, where $x(\cdot)\Big|_{[0,T]}$ stands for the restriction of the function $x(\cdot)$  to $[0,T]$. We shall denote by $\nabla g, \nabla^2g$ resp., the gradient and the Hessian of a function $g: \R^d \to \R$, assumed to be continuously differentiable, resp.\ twice continuously differentiable. A subscript to $\nabla$, when present, indicates the variable with respect to which the derivatives are evaluated. \\

\section{Preliminaries}

Consider the classical stochastic gradient descent to minimize a loss function $F: \R^d \to \R, \ d \geq 2,$ given by $F(x) = E[G(x, \xi)]$ where:
\begin{itemize}
\item $G: \R^d\times \R^{d'} \to \R^d$ is continuous and twice continuously differentiable in $x$ with bounded second partial derivatives, and,
\item $\xi$ is a $d$-dimensional random variable with distribution $\kappa$.
\end{itemize}
One has access to i.i.d.\ samples from $\kappa$. The classical SGD is the iterative scheme
\begin{equation}
x_{n+1} = x_n + a\left(-\nabla_x G(x_n, \xi_{n+1})\right), \ n \geq 0, \label{SGD0}
\end{equation}
where  $0 < a < 1$ is a fixed step size. Setting 
\begin{equation}
F(x) := E\left[G(x, \xi_n)\right], \ M_{n+1} = \nabla F(x_n) - E[\nabla G(x_n,\xi_{n+1}) | x_n], \label{Mdef}
\end{equation}
we have
\begin{equation}
x_{n+1} = x_n + a\left(-\nabla F(x_n) + M_{n+1}\right), \ n \geq 0, \label{SGD}
\end{equation}
where:
\begin{enumerate}

\item $F: \R^d \to \R$ is twice continuously differentiable with a Lipschitz gradient $\nabla F: \R^d \to \R^d$,

\item $M_n, n \geq 0,$ is a martingale difference sequence in $\R^d$, i.e., a sequence of square integrable random variables in $\R^d$ adapted to the increasing $\sigma$-fields
$$\F_n := \sigma(x(0); M_m, m \leq n), \ n \geq 0.$$
That is, $E\left[\|M_n\|^2\right] < \infty$ and
$$E\left[M_{n+1}|\F_n\right] = M_n \ \forall n \geq 0.$$
We impose the additional growth condition:
\begin{equation}
E\left[\|M_{n+1}\|^2 | \F_n\right] \leq K_1\left(1 + \|x_n\|^2\right) \ \forall n \geq 0. \label{mg}
\end{equation}
This implies a similar restriction on $G$. 
\end{enumerate}
The form \eqref{SGD}, while derived from \eqref{SGD0}, is in fact more general than \eqref{SGD0} and we shall conduct our analysis at this level of generality in this and the next section. We revert to \eqref{SGD0} in Section \ref{main} in order to further refine the results in the special context of \eqref{SGD0}.\\

We assume that the iterates are \textit{stable} in the sense that
\begin{equation}
\sup_nE\left[\|x_n\|^2\right] \ < \ \infty. \label{stable}
\end{equation}
This usually needs to be separately verified, see \cite{BorkarBook}, Chapter 4 for an overview of techniques for doing this. Under these assumptions, the ODE 
\begin{equation}
\dot{x}(t) = -\nabla F(x(t)) \label{ODE0}
\end{equation}
is well posed. The so called `ODE appproach to stochastic approximation' (SGD in the present case) goes back to \cite{Der}, \cite{Ljung}, \cite{Meer}. It views \eqref{SGD} as a noisy discretization of \eqref{ODE0}, leading to the conclusion that the iterates `track' the ODE in the following sense. For 
$$na \leq t < (n+1)a, \ m \geq 0,$$
let $\bar{x}(na) = x_n \ \forall n \geq 0,$ and $\bar{x}(t) :=$ the linear interpolation of $\bar{x}(ka) = x_k$ and $\bar{x}((k+1)a) = (k+1)a$ for $t \in [ka, (k+1)a]$, $k \geq 0$. Then $\bar{x}(\cdot)$ is a continuous, piecewise linear function $[0,\infty) \to \R^d$. The main result in this context is :  $\bar{x}(\cdot)$ `tracks' \eqref{ODE0} in the sense that for any $T > 0$,
\begin{equation}
\limsup_{t\uparrow\infty}\max_{s\in[t,t+T]}E\left[\|\bar{x}(s) - x(t)\|^2\right] = O(a), \ \mbox{a.s.} \label{const-step} 
\end{equation}
See, e.g., Chapter 9 of \cite{BorkarBook} and the references therein. Note that \eqref{const-step} implies concentration, not a.s.\ convergence, unlike the classical case of decreasing step sizes satisfying the Robbins-Monro conditions (see \cite{BorkarBook}, Chapter 2, for a pedagogical treatment of the latter). We stay with the constant step size formulation because that is the accepted practice in machine learning. See \cite{Azizian} for a comprehensive modern treatment of constant step size SGD using the ODE approach.\\

For notational ease, we write $h(\cdot) := -\nabla F(\cdot)$ and consider the differential equation in $\R^d$ given by
\begin{equation}
\dot{x}(t) = h(x(t)) \ (:= - \nabla F(x(t))), \ x(0) = x_0, \ t \geq 0. \label{ODE}
\end{equation}
where $h: \R^d \to \R^d$. By the Lipschitz continuity of $h = -\nabla F$,  $h$  satisfies the linear growth condition:
$$\|h(x)\| \leq K(1 + \|x\|)$$
which prevents finite time blow up (defined as $\lim_{t\uparrow T}\|x(t)\| = \infty$ for some $T < \infty$). This is easily verified using the Gronwall inequality and can be replaced by other conditions, e.g., existence of a suitable global Liapunov function. We take this as a given. Lipschitz continuity also ensures that the ODE \eqref{ODE} will be well posed, i.e., have a unique solution for any initial condition $x_0$ that depends continuously on $x_0$ as a map $\R^d \mapsto C([0,\infty);\R^d)$. (If $h$ is only continuous, one has existence, but uniqueness is not guaranteed \cite{Hartman}). \\

The focus of this article  is the case when $h$ is discontinuous across a manifold $\M$ embedded in $\R^d$, of dimension $m < d$. Furthermore, this is an important special case of `differential equations with discontinuous right hand side'. The latter have been extensively studied using differential inclusions, i.e., dynamics given by
\begin{equation}
\dot{x}(t) \in J(x(t)) \label{J}
\end{equation}
where $J$ is a set-valued map mapping $\R^d$ to convex compact subsets of $R^d$. There are multiple solution concepts for \eqref{J}, such as Krasovskii, Filippov and Hermes solutions, see, e.g.,  \cite{Cortes}, \cite{Filippov} and \cite{Hajek}. Our focus here is, however, for the special case mentioned above, where these notions are an overkill and one can say much more using the specific structure of the problem. \\

We shall analyze the passage from $\R^d$ to an embedded lower dimensional manifold. In case of intersecting manifolds, the passage from the dynamics on a manifold to that on a lower dimensional manifold is dealt with similarly. Also, our analysis is local, in a neighbourhood of a point on the manifold. See \cite{Durmus} for a recent work on constant step size stochastic approximation on a Riemannian manifold. 

\section{Vector fields discontinuous across a\\ manifold}

To fix ideas, recall the case of a co-dimension $1$ manifold $\M$ embedded in $\R^d$, analzyed in \cite{Filippov}. Let $x \in \M\cap O$ where $O \subset \R^d$ is an open ball that necessarily gets split into two parts with $\M\cap O$ as their common boundary. (This can be formalized using the \textit{tubular neighbourhood theorem}, see, e.g., \cite{Guil}, p.\ 76.) Without loss of generality, denote the two parts as $O^+, O^-$ respectively. Let $h^+(x), h^-(x)$ denote the drift at $x$ in $O^+, O^-$ resp. Write its normal and tangential components at $x$ as $h^\pm_n, h^\pm_t$ resp. By our hypothesis of transversality \textbf{$(\dagger)$}, $\|h_n^{\pm}(x)\| > 0$. Then the trajectory of the limiting differential equation at $x$ has a net drift of the form
\begin{equation}
g(x) = \alpha(x) h^+(x) + (1-\alpha(x))h^-(x) \label{g}
\end{equation}
where $\alpha(x) \in [0,1]$. Since the trajectory is confined to $\M$, the component of this vector field that is normal to $\M$ must cancel out. Thus we must have $\alpha h^+_n(x) = -(1-\alpha)h^-_n(x)$. That is,
$$\alpha = \alpha(x) = \frac{\|h^-_n(x)\|}{\|h^+_n(x)\| + \|h^-_n(x)\|}.$$
In particular, $\alpha(\cdot)$ and therefore $g(\cdot)$ inherits the regularity properties of $h(\cdot)$ near $\M$. Since the resulting vector field $g(x)$ is tangent to $\M$ at $x$, the normal component cancels and we may replace $h^{\pm}(x)$ in its definition by $h^{\pm}_t(x)$.  We thus have
\begin{eqnarray*}
g(x) &=& \left(\frac{\|h^-_n(x)\|}{\|h^+_n(x)\| + \|h^-_n(x)\|}\right)h^+_t(x) \\
&&\ + \  \left(\frac{\|h^+_n(x)\|}{\|h^+_n(x)\| + \|h^-_n(x)\|}\right)h^-_t(x) \\
&=&  \left(\frac{\|h^-_n(x)\|}{\|h^+_n(x)\| + \|h^-_n(x)\|}\right)h^+(x) \\
 && \ + \  \left(\frac{\|h^+_n(x)\|}{\|h^+_n(x)\| + \|h^-_n(x)\|}\right)h^-(x) \ ,
\end{eqnarray*}
where the second equality follows from (\ref{g}). In view of the Lipschitz continuity of $h$, the flow associated with the resulting differential equation along $\M$ is uniquely specified. See \cite{Filippov} for an extended exposition of the foregoing. An intuitive interpretation is that at any point $x\in\M$, the trajectory of (\ref{ODE}) is pushed towards $\M$ on both sides of $\M$ and the `relative frequency', so to say, of being on either side of $\M$ has to be such that it remains on $\M$. This implies that the frequency of approach to $\M$ on one side should be proportional to the magnitude of the normal component of the driving vector field on the other side.\\

We now formally extend this intuition to the case when $\M$ has dimension $m < d-1$. Let $\T_x, \N_x$ denote resp., the tangent and normal spaces at $x\in\M$. Their dimensions will be resp.\ $m$ and $d-m$. For a vector field $h(x)$ incident at $x\in\M$, let $h_t(x), h_n(x)$ denote its components in $\T_x, \N_x$, respectively. Then dim$(\N_x) \geq 2$. \\

Let $S_x$ denote the unit sphere in $\N_x$. 
Define an equivalence relation $\approx$ on $S_x$ by $s \approx -s, s \in S_x$. Let $S^*_x$ denote the corresponding quotient space, which is the familiar real projective space, a compact manifold. Then the equivalence class in $S_x^*$ that contains $s$ is simply $[s] := \{s,-s\}$. We can identify $s \in S_x$ with $s^* := ([s],\pm s)$, i.e., first identify the equivalence class and then the specific element thereof. 
Any probability measure $\mu \in \PA(S_x)$ can then be disintegrated as
$\mu(x,ds) = \mu_0(x,ds^*)\mu_1(s|x,s^*)$ where $\mu_0(x,\cdot) \in \PA(S^*_x)$ is the marginal on $S^*_x$ and $s^* \in S^*_x
\mapsto \mu_1( \cdot | x,s^*) \in \PA([s])$ is the regular conditional law on $[s] = \{s,-s\}$ given $s^*$, both parametrized by the location $x\in\M$. \\

Let $h^s(x)$ be a vector incident at $x$ along the direction $s\in S_x$, i.e., $\frac{h^s_n(x)}{\|h^s_n(x)\|} = s$. 
By the same argument as what was used in the codimension $1$ case, the net vector field of $h^s(x)$ and $h^{-s}(x)$ along $\M$ at $x$ is given by
\begin{eqnarray*}
&&\frac{\|h^{-s}_n(x)\|h^s(x) + \|h^{s}_n(x)\|h^{-s}(x)}{\|h^{-s}_n(x)\| + \|h^s_n(x)\|} \\
&=& \frac{\|h^{-s}_n(x)\|h^s_t(x) + \|h^{s}_n(x)\|h^{-s}_t(x)}{\|h^{-s}_n(x)\| + \|h^s_n(x)\|}\ .
\end{eqnarray*}
In the notation introduced above, this translates into
\begin{eqnarray*}
\mu_1(s|x,s^*) &=& \frac{\|h^{-s}_n(x)\|}{\|h^{-s}_n(x)\| + \|h^s_n(x)\|}, \\
 \mu_1(-s|x,s^*) &=& \frac{\|h^{s}_n(x)\|}{\|h^{-s}_n(x)\| + \|h^s_n(x)\|}.
 \end{eqnarray*}
Thus the effective vector field driving the dynamics along $\M$ at $x$ is given by
$$h^*(x) := \int\left(\frac{\|h^{-s}_n(x)\|h^s(x) + \|h^{s}_n(x)\|h^{-s}(x)}{\|h^{-s}_n(x)\| + \|h^s_n(x)\|}\right)\mu_0(x,ds^*).$$
The effective o.d.e.\ in $\M$ is then given by
\begin{equation}
\dot{x}(t) = h^*(x(t)). \label{main}
\end{equation}
Our next task is to pin down the measure $\mu_0$, which we call the `averaging measure'. \\

\section{Main result}\label{main}

We now adapt a paradigm from \cite{Vark} to our framework. The dynamics \eqref{main} characterizes the effective flow along $\M$. Now consider the component flow along $\SA_x$ in $O$. Let $r > 0$ denote the radius of $O$. Let $h^s_p(x) :=$ the component of $h(x)$ along $\T_x$ at $s \in \M$. This defines a gradient flow along $S_x$. Then by the same logic as above, the effective vector field driving the flow on $S^*_x$ is 
$$\tilde{h}(x) := \frac{\|h^{-s}_n(x)\|h^s_p(x) + \|h^{s}_n(x)\|h^{-s}_p(x)}{\|h^{-s}_p(x)\| + \|h^s_p(x)\|}\ .$$
Taking a cue from \cite{Vark}, we make the following important assumption:\\

\noindent \textbf{Assumption A1:}\label{VA} The limit $h^\dagger : S^*_x \to \R^d$ defined as $h^\dagger(x) := \lim_{0 < r \downarrow 0}\frac{\tilde{h}(rx)}{r}$ is well defined and Lipschitz in the local coordinates of $S^*$.\\

Denote the limiting flow on $S^*_x$ by 
\begin{equation}
\dot{x}(t) = h^\dagger(x(t)) \label{limit}
\end{equation}
in the limit $0 < r \downarrow 0$. This is a gradient flow in the compact manifold $S^*_x$. Hence it will converge to the set of critical points where the gradient vanishes. Following a standard rationale  used in the analysis of stochastic approximation algorithms, we ignore the unstable critical points, that is, local maxima, saddle points and points of inflection, by invoking `avoidance of traps' results that establish a.s.\ non-convergence of the algorithm to unstable equilibria. This class of results gives conditions on the martingale difference noise $\{M(n)\}$ for the foregoing to be true\footnote{essentially, that it be `rich enough' in all directions in a suitable sense, the exact formulation of this varies across different articles}, see \cite{BorkarBook} Section 3.4 and the references therein. In particular, this allows us to focus only on local minima. We further assume that the set of these local minima are finitely many and isolated. \\

We now consider a specific, albeit the most important special case of the SGD introduced above in \eqref{Mdef}. Under \eqref{stable}, $\{x_n\}$ in \eqref{SGD0} is a stable Markov process. Hence it will have a non-empty convex compact set of invariant probability measures, the extreme points of which correspond to ergodic processes. (See, e.g., \cite{MT}.) We shall assume that the invariant measure is unique and write it as $\nu_a$ in order to render explicit its dependence on the step size $a$. From \cite{Azizian}, we know that $\nu_a$ satisfies a large deviations principle \cite{Varadhan} with an associated potential function $V$, which may in general differ from $F$ (see \textit{ibid.}). It follows that as $a \downarrow 0$,  Then the results of \cite{Hwang} lead to the following. We assume that the global minima $\{x_i\}$ of $F$ are not flat in any direction, i.e., the Hessians $\{DV_{x_i}\}$ are non-singular.

\begin{theorem} As $0 < a\downarrow 0$, $\mu_a$ concentrates on the global minima $x_1, \cdots , x_\ell$ (say) of $F$, with weights proportional to det$\left(DV_{x_i}\right)$ for $1 \leq i \leq \ell$. \end{theorem}

Let $\delta(x)(dy)$ denote the Dirac measure at $x \in \R^d$. Then our main result is as follows.

\begin{theorem} The measure $\mu(x,dy) \in \PA(S_x)$ is given by 
$$\mu(x,dy) \ = \  C\sum_{i=1}^\ell \mbox{det}\left(DV_{x_i}\right)\delta_{[x_i]}(dy),$$ where $C > 0$ is the normalizing factor given by
$$C := \left(\sum_{i=1}^\ell \mbox{det}\left(DV_{x_i}\right)\right).$$
\end{theorem}

This is a direct consequence of Theorem 2.1 of \cite{Hwang}. In \cite{Hwang}, the basic result is extended to the case when the set of global minima forms a union of finitely many embedded manifolds $\N_i, 1 \leq i \leq M$ (say). It is assumed that $V$ is thrice continuously differentiable. A base probability measure $Q$ on $\R^d$ is given, with continuous density $f(\cdot) > 0$ with respect to the Lebesgue measure. The notion of `\textit{intrinsic measures}' on the highest dimensional manifolds is defined, see \textit{ubid.} for details. Denote by $\zeta$ the sum of these measures. The result (see Theorem 3.1 of \cite{Hwang}) then is as follows.

\begin{theorem} As $0 < a\downarrow 0$, $\mu_a \to \mu$ in $\PA(S_x)$, where $\mu$ is supported on the union of the $\N_i$'s with the highest dimension and is given by
$$\mu(x,dz) \ = \ \left(\frac{f(z)\mbox{det}\left(DV_z^{-1/2}\right)}{\int \left(f(z')\mbox{det}\left(DV_z\right)\right)dz'}\right) \zeta(dz) .$$
\end{theorem}

\section{The time evolution}

The above analysis is at a single point in the embedded manifold $\M$. As time evolves, the set of global minima will also evolve, therefore  so will the above measure-valued process supported on this set. If the condition det$DV_y \neq 0$ holds at a point $y$, it will hold in a neighbourhood thereof by continuity, hence the points in the ODE trajectory where it holds form a union of disjoint open intervals separated by points $y$ where det$DV_y = 0$. The latter corresponds to points where some global minima merge and/or split. For simplicity, suppose the number of global minima remains bounded from above. We can label them lexicographically according to their positions. Then the possible   merging and/or splitting patterns for global minima can also be enumerated in terms of this labelling.  We can map these into a finite alphabet, say $\LA$. Augment $\LA$ by adding a distinct symbol, say  $\upsilon$, for no merging or splitting. Consider the function $\Psi: [0,\infty) \to \LA$ that maps $x(t)$ in \eqref{limit}  to its symbol in $\LA$. Then 
$$\Psi^{-1}(\upsilon) = \{t \geq 0 : \mbox{det}\left(DV_{x(t)}\right) \neq \0\}$$ 
will be a union of open intervals.  On the other hand, distinct splitting and/or merging events cannot be concurrent and therefore 
$$H := \Psi^{-1}(\LA\backslash\{\upsilon\}) = \{t \geq 0 : \mbox{det}\left(DV_{x(t)}\right) = \0\}$$
cannot contain an interval. However, it appears that the possibility of the points in $H$ not  being isolated (i.e., being an accumulation point) cannot be ruled out. Nevertheless, in practice, it is reasonable to expect that this does not happen. This given a complete picture of the dynamic evolution of the SGD on $\M$.

\end{document}